# EfficientCD: A New Strategy For Change Detection Based With Bi-temporal Layers Exchanged

Sijun Dong, Yuwei Zhu, Geng Chen, Xiaoliang Meng*

**Abstract—With the widespread application of remote sensing technology in environmental monitoring, the demand for efficient and accurate remote sensing image change detection (CD) for natural environments is growing. We propose a novel deep learning framework named EfficientCD, specifically designed for remote sensing image change detection. The framework employs EfficientNet as its backbone network for feature extraction. To enhance the information exchange between bi-temporal image feature maps, we have designed a new Feature Pyramid Network module targeted at remote sensing change detection, named ChangeFPN. Additionally, to make full use of the multi-level feature maps in the decoding stage, we have developed a layer-by-layer feature upsampling module combined with Euclidean distance to improve feature fusion and reconstruction during the decoding stage. The EfficientCD has been experimentally validated on four remote sensing datasets: LEVIR-CD, SYSU-CD, CLCD, and WHUCD. The experimental results demonstrate that EfficientCD exhibits outstanding performance in change detection accuracy. The code and pretrained models will be released at https://github.com/dyzy41/mmrscd.**

## I. INTRODUCTION

REMOTE sensing image change detection plays a pivotal role in the field of environmental monitoring and urban development and water conservancy. Given the rapid changes in the Earth's environment, such as urban expansion, natural disasters, and ecological degradation, accurately and efficiently monitoring these changes is crucial for environmental protection, urban planning, and disaster prevention and response. Remote sensing technology, by capturing surface images at different time points, provides reliable visual evidence of these changes, making remote sensing image change detection feasible. In recent years, the application of deep learning technology in remote sensing image processing has made remarkable progress. By leveraging the powerful feature-fitting capabilities of deep learning models and high-performance computing equipment, the accuracy and efficiency of remote sensing image change detection have been significantly enhanced [1], [2], [3]. These deep learning models are capable of processing large-scale remote sensing data, identifying and analyzing subtle surface changes, thereby playing an increasingly important role in environmental monitoring.

In the field of deep learning for remote sensing image change detection, many researchers have optimized various aspects of the process [4], [5]. The core objective of change detection algorithms is to detect regions within bi-temporal images where changes have occurred. The model needs to identify the changed areas between the two images taken at different times. To achieve this goal, we can focus on two main areas: first, enhancing the model's ability to extract features from bi-temporal images, thereby improving its representation of remote sensing images. Second, designing more sophisticated modules to express differential features, making it easier for the change detection model to identify changed areas in bi-temporal images.

Specifically, in terms of feature extraction backbones, some researchers have chosen the advanced Vision Transformer [6] for feature extraction of bi-temporal images. Through the powerful feature extraction capabilities of these backbones, the model's performance in change detection tasks has been improved [7], [8], [9]. Meanwhile, considering the unique characteristics of change detection, many researchers have focused on the computation of differential features [1], [10], [11]. Accordingly, the aforementioned two types of research optimize the change detection task from different perspectives. The former enhances the model's ability to understand different land cover targets within remote sensing images, thereby allowing the model to better recognize various land cover categories in bi-temporal images. The latter, on the other hand, enhances the model's ability to perceive changed features which is the foundation of change detection tasks.

In common change detection algorithms, methods such as concatenation, addition, and subtraction are typically used to fuse the bi-temporal features to represent the differential features. From a computational perspective, feature concatenation does not directly reflect the differences in bi-temporal feature maps. It mainly learns the changed areas through the strong fitting relationship between the model and

Accepted in IEEE Transactions on Geoscience and Remote Sensing in 20-Jul-2024.

This work was funded by Key Research and Development Plan of Guangxi Zhuang Autonomous Region: 2023AB26007 and the National Natural Science Foundation of China (NSFC): 41971352. (Corresponding author: Xiaoliang Meng.)

Sijun Dong, yuwei Zhu and Xiaoliang Meng are with the School of Remote Sensing and Information Engineering, Wuhan University, Wuhan 430079,

China (e-mail: dyzy41@whu.edu.cn; yuweizhu@whu.edu.cn; xmeng@whu.edu.cn).

Xiaoliang Meng is with Hubei LuoJia Laboratory, Wuhan University, 430079 Wuhan, China.

Geng Chen is with the Guangxi Water& Power Design Institute CO., Ltd.Minzhu road 1-5, Nanning, Guangxi, 530027, China.(e-mal:chengenghhu@163.com)

the change labels [12]. Addition retains the foreground areas of bi-temporal images, but the model needs to learn which foreground areas represent the changed regions [11]. Subtraction eliminates the influence of identical elements in bi-temporal images [13]. However, the subtraction method increases the numerical difference within changed features, making it more challenging for the model to distinguish the changed areas. Therefore, designing a suitable method to better construct the representation of differential features has always been a research hotspot in the field of change detection.

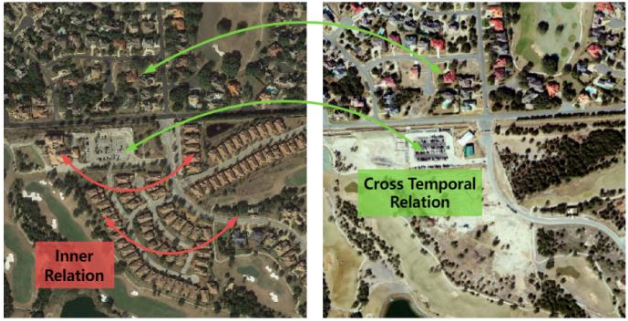

**Fig. 1.** The pixel relationship in Change Detection

In the field of semantic segmentation, particularly in remote sensing image segmentation, the fusion of features from different levels has significantly enhanced model performance [14], [15], [16]. These feature fusion strategies, by integrating information from various depths of the network, effectively enhances the model's ability to recognize targets of different scales and complexities within images [17]. In remote sensing image segmentation, this method improves the handling of detail changes in images taken from high altitudes, such as small-scale surface features and complex terrain structures, thereby achieving more precise land cover classification [18] and change detection [2]. For change detection tasks, the unique bi-temporal images provide additional optimization directions for multi-level feature fusion [2], [19], [20]. We notice a significant similarity in land cover categories between bi-temporal images, as illustrated in Figure 1. Therefore, it is beneficial to build pixel dependencies both within single-temporal images and between bi-temporal images to enhance the representation capability of the models.

Unlike the studies mentioned previously, this paper devises an optimization scheme for change detection algorithms focusing on the enhancement of intra-temporal image features and the interaction of features between bi-temporal images. Specifically, the main contributions of this paper are as follows:

(1) Based on the structural characteristics of EfficientNet, this paper constructs a multi-level feature pyramid suitable for remote sensing image change detection tasks.

(2) To strengthen the feature interaction between bi-temporal remote sensing images, this paper designs the ChangeFPN architecture, which improves feature communication between bi-temporal images through a parameterless approach.

(3) This paper designs a layer-by-layer decoding architecture for the change detection task based on the Euclidean distance of the feature map.

## II. RELATED WORKS

### A. Feature Extraction Backbone in Change Detection

As deep learning technology continues to evolve, a multitude of backbone models dedicated to feature extraction have emerged, such as ResNet [21], ResNeSt [22], Vision Transformer [6], and EfficientNet [23]. Consequently, many researchers have designed change detection algorithms based on these novel backbone networks. Yuan et al. introduced the STransUNet model [24], which is the first to integrate the advantages of UNet, CNN, and Transformer [25] in change detection tasks. The CNN models extract detailed information from low-level features, while the Transformer models focus on long-term global attention in high-level features. Compared to solely using the Transformer structure to facilitate the optimization of change detection tasks, the backbone network based on the Swin Transformer [26] model surpasses the Vision Transformer [6] in terms of feature extraction. As a result, some researchers have shifted their focus to using the Swin Transformer to optimize change detection tasks. Zhang et al. proposed SwinSUNet [8], the first pure transformer network for change detection tasks. Compared with previous CNN-based methods, SwinSUNet excels at extracting spatiotemporal global information. In contrast to the Vision Transformer structure based on transformer modules, the EfficientNet structure proposed by Tan et al. [23] offers a better balance between model accuracy and efficiency. Both Siam-EMNet [27] and EffCDNet [9] utilize bi-branch pretrained EfficientNet models as the encoding structure to extract features from bi-temporal images. The superior accuracy and better computational efficiency of the EfficientNet model compared to complex transformer structures make it preferable to traditional CNN architectures for change detection networks.

In addition, many researchers today are dedicated to foundational model research in remote sensing vision [28], [29]. These foundational model architectures [30], [31] have better generalization capabilities and recognition performance for remote sensing images compared to traditional CNN architectures. Sun et al. leverage the benefits of generative self-supervised learning (SSL) for remote sensing (RS) images and propose a foundational model framework for RS called RingMo [32]. They also propose a foundational model training method specifically for remote sensing data and large-scale datasets. In terms of multi-modal remote sensing images, Guo et al. present SkySense, a generic billion-scale model pre-trained on a curated multi-modal Remote Sensing Imagery (RSI) dataset with 21.5 million temporal sequences [33]. Liu et al. construct a large-scale remote sensing image-text dataset based on the CLIP model and build a Vision Language Foundation Model for Remote Sensing through contrastive learning named RemoteCLIP [34]. This introduces more exploration into combining textual knowledge in the field of remote sensing. To explore the application of foundational models in the remote sensing field, Dong et al. first propose a change detection network based on the CLIP model that combines vision and language representation learning [1]. Li et al. propose BAN to utilize general knowledge from large foundational models to reduce the dependence of existing change detection (CD) models on large amounts of labeled data [35]. The application of these foundational models in the field of remote sensing typically involves training foundational models with ultra-large-scale data to enhance the generalization capability of the backbone in remote sensing image recognition, thereby improving the models' performance in various downstream tasks. The construction of large foundational models usually requires significant computational resources. Additionally, due to the high diversity of remote sensing sensors, remote sensing foundational models may not necessarily be applicable to all scenarios.

Unlike the methods mentioned above, the EfficientCD mainly focuses on the model architecture of the change detection algorithm. In fact, the architecture of EfficientCD can be easily combined with the aforementioned foundational models. Based on the hierarchical design scheme of the EfficientNet model, we extracted more feature layers to construct the feature pyramid of the change detection model, thereby obtaining a richer array of meaningful features during the feature extraction process. On one hand, we leverage the powerful feature extraction capability and computational efficiency of EfficientNet. On the other hand, the multiple levels of feature maps extracted from the EfficientNet structure better match the characteristics of remote sensing imagery. This is because multiple levels of feature maps can fully capture the scale differences of objects in remote sensing images.

*B. Feature Interaction*

Both are intensive prediction tasks; however, change detection tasks differ from image semantic segmentation by having bi-temporal input data. In change detection tasks, the bi-temporal inputs share the same geographical location but differ in time phases. Therefore, due to the difference in time phases, the bi-temporal input data may exhibit differences in land cover categories. However, since the geographical locations are the same, the land cover categories in the bi-temporal images will be correlated. Thus, images from different time phases can enhance the learning ability of the change detection model through the construction of information interaction modules.

Specifically, Liu et al. proposed the FADL module [36] to enable the model to learn change areas in bi-temporal images. At the level of bi-temporal feature interaction, the authors refined the bi-temporal features through the JFR module, enhancing features of the change areas by increasing attention to spatial and channel dimensions. Wang et al. introduced the BFDEB module [37], designed to enhance post-event image features and bi-temporal image difference features. These modules ensure clear boundaries and consistent internal logic in the results, reducing false detections and missed detections caused by interference factors. Gu et al. [11] proposed the DDFM module, designed to mine and fuse multiple differences at each feature level between bi-temporal images, improving the completeness of the detected changed objects and the ability to alleviate pseudo-changes. These approaches mainly study how to fuse features of bi-temporal images, thereby enabling the network model to better express the features of change areas.

Unlike the methods mentioned above, the ChangerEx [38] model partially exchanges the features of bi-temporal images during the feature extraction process. ChangerEx employs two types of exchange strategies: channel exchange and spatial exchange. Channel exchange refers to exchanging features in the channel dimension, while spatial exchange refers to exchanging features in the spatial dimension. Typically in the deep learning domain, the channel dimension of feature maps represents the semantic information of images, while the spatial dimension represents the spatial texture information of images. Therefore, in other computer vision fields, channel and spatial exchanges typically disrupt the original image information. However, for change detection, the goal is to detect changes occurring in bi-temporal images. Thus, any interactive change between bi-temporal features does not negatively affect the computation of feature differences. Meanwhile, feature exchange strengthens the information communication of bi-temporal features to some extent, similar to constructing the pixel dependency of bi-temporal images. The construction of such pixel dependency can enhance the characterization ability of the model [39].

Here, according to the feature interaction method in change detection, we call the above feature interaction method for the corresponding feature map a feature-based method. Differing from the aforementioned feature interaction methods, we propose a new feature interaction approach called ChangeFPN. ChangeFPN exchanges corresponding feature levels of the feature pyramid through an alternating selection method. We call it a layer-based method. Without adding any additional parameters, ChangeFPN enhances information interaction between features of bi-temporal images, thereby strengthening the representational capability of the change detection model. Because ChangeFPN has a different starting point from the aforementioned methods based on bi-temporal feature maps, it can be integrated with commonly used feature interaction algorithms for feature maps, thereby effectively enhancing change detection capabilities.

## III. METHOD

*A. Overall Architecture*

In this paper, we have designed a change detection network based on the EfficientNet backbone, as illustrated in Figure 2. The figure displays a change detection model based on the EfficientNet-B5 backbone network. The proposed EfficientCD consists of three parts: a feature extraction backbone network based on EfficientNet, the ChangeFPN for facilitating information exchange between different levels of bi-temporal images, and a layer-by-layer decoding module. The backbone leverages the EfficientNet architecture to extract rich and efficient features from the input images, and the dual-encoder shares the same weights. This backbone is chosen for its balance of accuracy and efficiency, making it particularly suitable for the demands of change detection tasks, which require processing large and complex datasets. The ChangeFPN module is designed to enhance the interaction between features extracted from bi-temporal images at different levels, as shown in Figure 3. By allowing the exchange of information between corresponding levels of the feature pyramid, ChangeFPN aims to improve the model's ability to detect and highlight changes between bi-temporal images. Finally, the multi-level decoding module progressively reconstructs the final change detection map from the combined pyramid feature maps. This layer-by-layer approach ensures the model's ability to understand target changes at different scales in remote sensing images. In the decoding stage, we use the yellow feature maps to calculate multiple auxiliary losses and the red feature maps to calculate the primary loss. The inference results are upsampled from the red feature map.

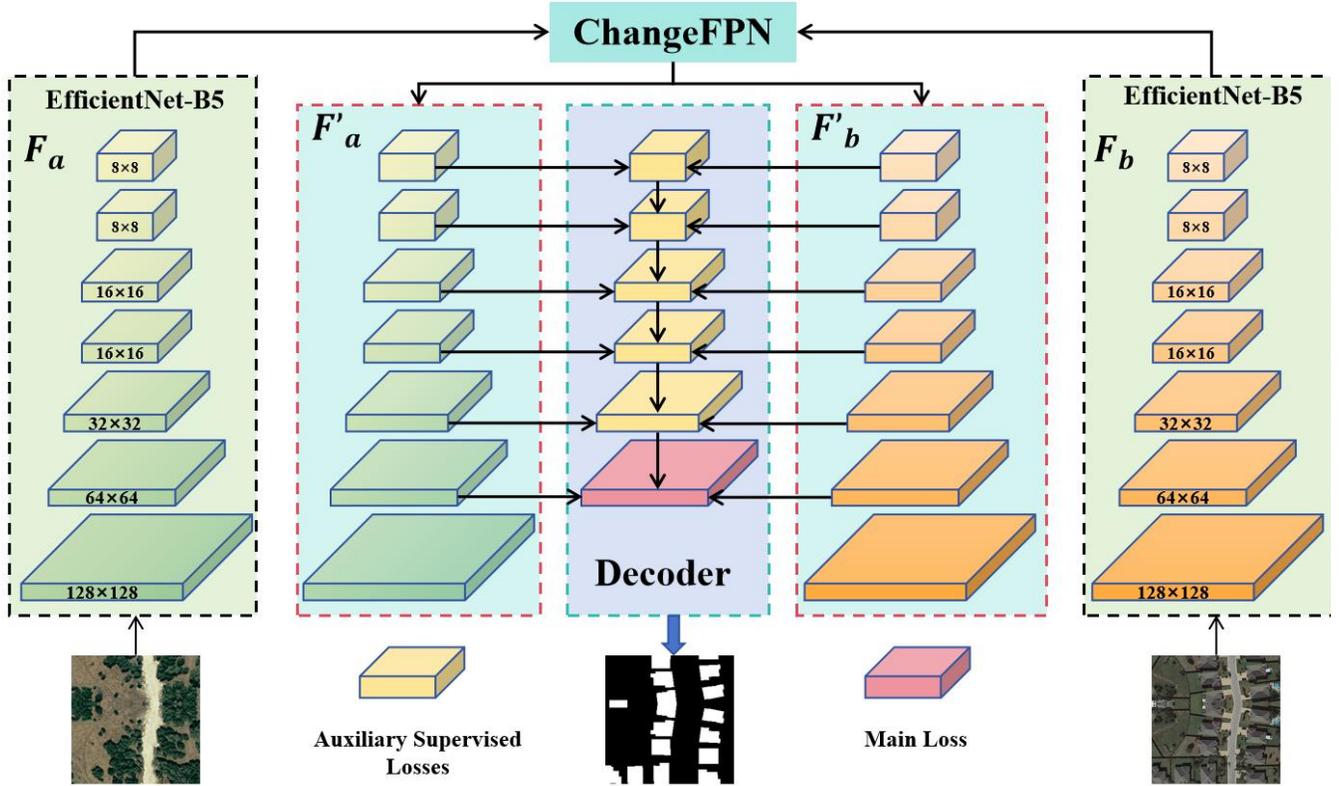

**Fig. 2.** The overall architecture of EfficientCD

*B. EfficientNet*

Most current change detection algorithms revolve around the vision transformer architecture. The full-pixel computation method of the self-attention module in the vision transformer architecture endows feature extraction backbones based on vision transformers with a global cognitive ability for remote sensing images. Consequently, change detection tasks utilizing the vision transformer architecture have achieved significant success across various datasets. However, typically, due to the extensive high-dimensional matrix multiplication calculations in the self-attention module, vision transformers have high computational costs. This high computational demand also suggests that the model has a better fitting ability for the features of remote sensing images.

In contrast to the approaches mentioned above, this paper opts for the lightweight network structure EfficientNet as the feature extraction backbone for change detection tasks. For conventional convolutional neural networks (CNNs), enhancing the model's fitting capability can be achieved by increasing the depth and width of the network. However, deepening and widening the network also leads to increased computational consumption and may result in more severe overfitting issues during network training. Change detection tasks, commonly aimed at binary classification, focus solely on detecting the change areas of foreground targets. Additionally, compared to natural scene images, remote sensing images contain fewer specific semantic category targets. Therefore, considering the use of EfficientNet as the feature extraction backbone for change detection tasks is deemed appropriate. To this end, based on the structural characteristics of EfficientNet, we designed the feature extraction backbone as shown in Figure 2. We chose EfficientNet-B5 as the main backbone network, with its architectural structure presented in TABLE I.

TABLE I
EfficientNet-B5 Architecture

| Stage | Operator | Kernel size | Resolution | Channels | Layers |
|---|---|---|---|---|---|
| 1 | Conv | 3×3 | 128×128 | 48 | 1 |
| 2 | MBConv1 | 3×3 | 128×128 | 24 | 3 |
| 3 | MBConv6 | 3×3 | 64×64 | 40 | 5 |
| 4 | MBConv6 | 5×5 | 32×32 | 64 | 5 |
| 5 | MBConv6 | 3×3 | 16×16 | 128 | 7 |
| 6 | MBConv6 | 5×5 | 16×16 | 176 | 7 |
| 7 | MBConv6 | 5×5 | 8×8 | 304 | 9 |
| 8 | MBConv6 | 3×3 | 8×8 | 512 | 3 |

The MBConv mentioned in TABLE I refers to a module within the architecture, where the number following MBConv (such as 1 or 6) represents the expansion factor. This means that the first 1x1 convolution layer within the MBConv module will expand the number of channels in the input feature matrix by a factor of n. Kernel size denotes the size of the convolution kernel used by the Depthwise Convolution [40] within the MBConv module. Resolution indicates the input resolution for that stage. 'Channels' refers to the number of channels in the output feature matrix after passing through that stage. 'Layers' signifies how many times the operator structure is repeated within that stage.

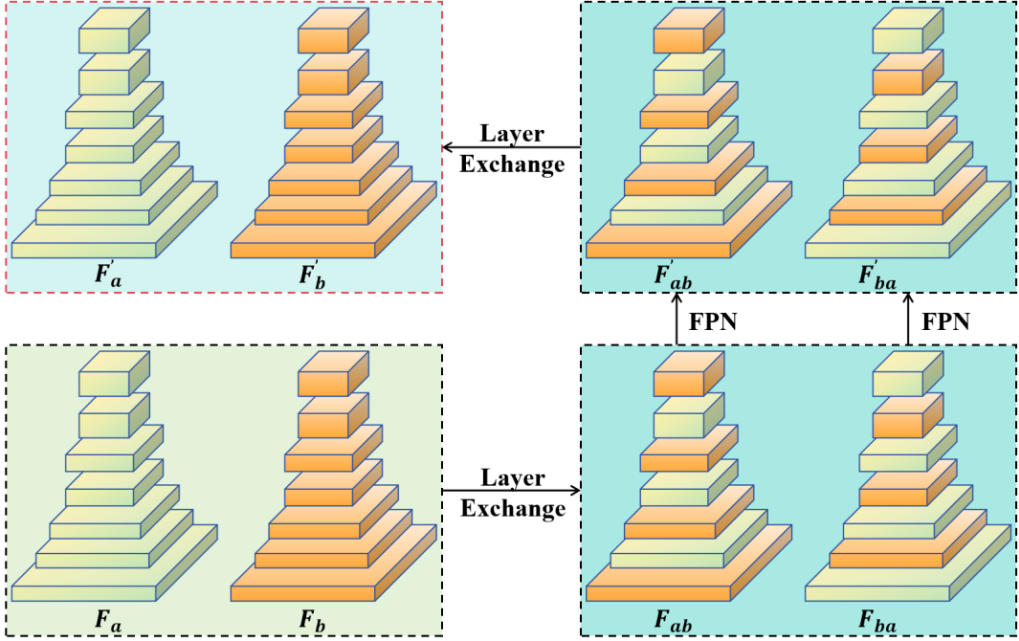

**Fig. 3.** The feature exchange method in ChangeFPN.

### C. ChangeFPN

In the field of computer vision, the fusion of features at different levels enhances the model's ability to perceive targets of varying scales. However, unlike semantic segmentation tasks, the geographical location remains the same for bi-temporal images in change detection, with differences in the images' land cover elements due to temporal variations. Nevertheless, given the consistency in geographical location, there should be a strong correlation between the bi-temporal images. To enable the model to learn more correlations between bi-temporal images, we designed the ChangeFPN structure. As shown in Figure 3 above, $F_a$ and $F_b$ represent the feature pyramids of bi-temporal images and $F'_a$ and $F'_b$ represent the output of ChangeFPN, same as the Figure 2. To ensure the thorough integration of information between the bi-temporal images, we opted to exchange adjacent layer features of the bi-temporal feature pyramids. $F_{ab}$ and $F_{ba}$ represent the results after the exchange. We applied the FPN model for $F_{ab}$ and $F_{ba}$, generating $F'_{ab}$ and $F'_{ba}$. Among them, $F_{ab}$ and $F_{ba}$ have feature layers with each other, and we perform multi-level feature fusion through FPN on the $F_{ab}$ and $F_{ba}$. The resulting output feature pyramid $F'_{ab}$ and $F'_{ba}$ integrate the features of the bi-temporal images. Through this computation method, we not only strengthened the information fusion at different levels but also constructed the information fusion between bi-temporal image feature maps.

To ensure that the method of feature exchange does not affect the subsequent decoding part, we used a layer exchange calculation for restoration. Unlike other feature-based methods that construct interactions between bi-temporal image features, our layer-based method increases the information exchange between bi-temporal images in a clever way, which improves the change detection model's ability to perceive change characteristics without increasing the computational load.

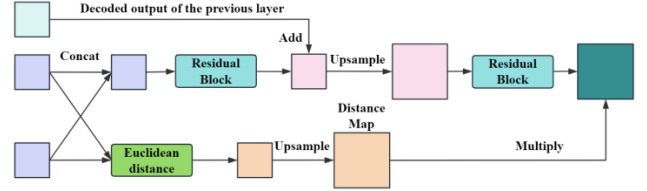

**Fig. 4.** Bi-temporal feature fusion module combined with Euclidean distance

### D. Layer-By-Layer Decoder Combined With Euclidean Distance

In the decoding stage of EfficientCD, we adopted a hierarchical approach to process the features outputted by ChangeFPN and calculated the change between the bi-temporal feature maps at each level using the Euclidean distance, as shown in Figure 4. For each layer of the bi-temporal feature maps from the FPN module, we first concatenate the bi-temporal feature maps (labeled as $F_1$ and $F_2$) along the channel dimension C to create a merged feature map. Subsequently, we calculate the Euclidean distance between these two feature maps to estimate the degree of difference between the bi-temporal feature maps. The formula for calculating the Euclidean distance is as follows:

$$D = \sqrt{\sum_C (F_1 - F_2)^2} \tag{1}$$

$$d = max(D) \tag{2}$$

$$D_{\text{norm}} = \sigma\left(\frac{d}{D_{max}}\right) \tag{3}$$

In the formula (1-3), $D$ represents the Euclidean distance matrix between $F_1$ and $F_2$, $\sigma$ denotes the sigmoid function, and the $D_{norm}$ is normalized. As shown in Figure 4, the two purple modules on the left represent the bi-temporal feature maps $F_1$ and $F_2$. First, the bi-temporal feature maps $F_1$ and $F_2$ are concatenated to generate a fused feature map. The fused feature map is then processed by a residual block, while simultaneously adding the decoded feature map from the previous layer, thus constructing a layer-by-layer feature fusion during the decoding process. This way method multi-scale information during the decoding stage. The fused feature map then undergoes bilinear interpolation upsampling to restore its spatial resolution. The upsampled feature map is further processed by another residual block. To enhance the model's sensitivity to change regions, the bi-temporal feature maps $F_1$ and $F_2$ are processed by the Euclidean distance module to calculate a distance map, as shown in equations (1-3). The distance map also undergoes bilinear interpolation upsampling to match the resolution of the feature map. Then, the upsampled distance map and the feature map are multiplied element-wise, enhancing the features in the regions with significant changes. These steps constitute a decoding module that incorporates Euclidean distance, effectively improving the model's ability to detect changes through layer-by-layer decoding and feature fusion. This layer-by-layer decoding method is more suitable for the decoding part of remote sensing image dense prediction tasks, because remote sensing image recognition usually requires combining different levels of feature information to achieve better recognition results.

Additionally, we implemented a multi-level supervision strategy, selecting multiple layers of feature maps to compute the loss. The first five feature maps in Figure 2 are used to compute auxiliary loss, while the sixth feature map is used to compute the main loss for the final prediction results. The inference results are also generated from the sixth feature map. This paper adopts cross-entropy loss as the loss function, with its formula as follows：

$$L = -\frac{1}{N}\sum_{i=1}^{N}\left(y_i\log\left(\hat{y}_i\right) + (1-y_i)\log\left(1-\hat{y}_i\right)\right) \quad (4)$$

$N$ is the total number of pixels in the image. $y_i$ is the real label of the pixel $i$. $\hat{y}_i$ is the probability that the model predicts that the pixel $i$ belongs to category 1. Meanwhile, to build multi-level auxiliary loss supervision, we selected feature maps at multiple levels in the decoding process for auxiliary loss supervision. The total loss function is calculated as follows:

$$L_{total} = L_{main} + \sum_{k=1}^{K} (1.00/K) * L_{aux_k} \quad (5)$$

$L_{main}$ is the main loss, usually the cross-entropy loss calculated for the prediction results of the output layer. $K$ is the number of auxiliary losses, corresponding to the number of feature maps which is used as the auxiliary supervision layers. $L_{aux_k}$ is the auxiliary loss of the k-th level, usually also the cross-entropy loss, and calculates the prediction results corresponding to the feature maps of this level.

## IV. EXPERIMENTS

### A. Datasets

In this study, we leveraged four sub-meter resolution datasets—LEVIR-CD [41], WHUCD [42], SYSU-CD [43], and CLCD [44]—to showcase the robustness and versatility of our change detection algorithm across diverse environments and scenarios. The following is a detailed introduction to these datasets.

The LEVIR-CD dataset, introduced by Chen et al., serves as a pivotal resource for conducting building change detection. It encompasses 637 pairs of very high-resolution image slices from Google Earth, each with a resolution of 0.5 m/pixel, and dimensions of 1024×1024 pixels. This dataset is specifically tailored to capture alterations within building areas, with precise annotations for both changed and unchanged building regions. It boasts a comprehensive collection of 31,333 instances of individual building changes. For algorithm training and evaluation, the dataset is divided into sets for training, validation, and testing, comprising 445, 64, and 128 image pairs respectively. Adhering to standard experimental protocols, these image pairs are further segmented into smaller patches of 256×256 pixels, maintaining a 64-pixel overlap between patches. For areas where the dimensions cannot be aligned, we opted to use mirror padding on the original images to ensure they can be precisely cut. This segmentation process yields 16,020 training, 2,304 validation, and 4,608 test image patches, respectively.

The WHUCD dataset consists of two aerial images with a resolution of 32507×15354 pixels and a spatial resolution of 0.3 meters per pixel. Similar to the LEVIR-CD dataset, the WHUCD dataset is specifically designed for building change detection. It has been widely used in various studies as a benchmark dataset for evaluating change detection algorithms. However, there is no standardized data split method for the WHUCD dataset. To ensure fair comparison and reproducibility, we adopted a consistent data partition scheme and replicated several existing algorithms for comparison against our proposed EfficientCD. Following the works in [41], [42], [45], we divided the original images into non-overlapping chips of 256×256 pixels. These chips were randomly split into three subsets: 6096 chips for training, 762 chips for validation, and 762 chips for testing, following the same splitting strategy as in previous works.

The SYSU-CD dataset is a collection of 20,000 pairs of high-resolution aerial images of 256×256 pixels captured in Hong Kong between 2007 and 2014. The dataset contains various types of changes, including the construction of new urban buildings, suburban sprawl, pre-construction groundwork, vegetation changes, road extensions, and offshore construction. The dataset is divided into three sets: the training set, validation set, and test set, consisting of 12,000, 4,000, and 4,000 image pairs, respectively.

The CLCD dataset consists of 600 pairs of farmland change sample images, with 320 pairs are used for training, 120 pairs are used for validation, and 120 pairs are used for testing. The bi-temporal images in the CLCD were collected by Gaofen-2 in Guangdong Province, China, in 2017 and 2019, respectively, with spatial resolutions ranging from 0.5 to 2 meters. Each set of samples consists of two 512×512 images and a corresponding binary label for farmland changes. During the training process, we randomly cut the image into 256×256 size for training. During the prediction inference process, we use sliding window prediction into 256×256 size, and the sliding window size is 170×170.

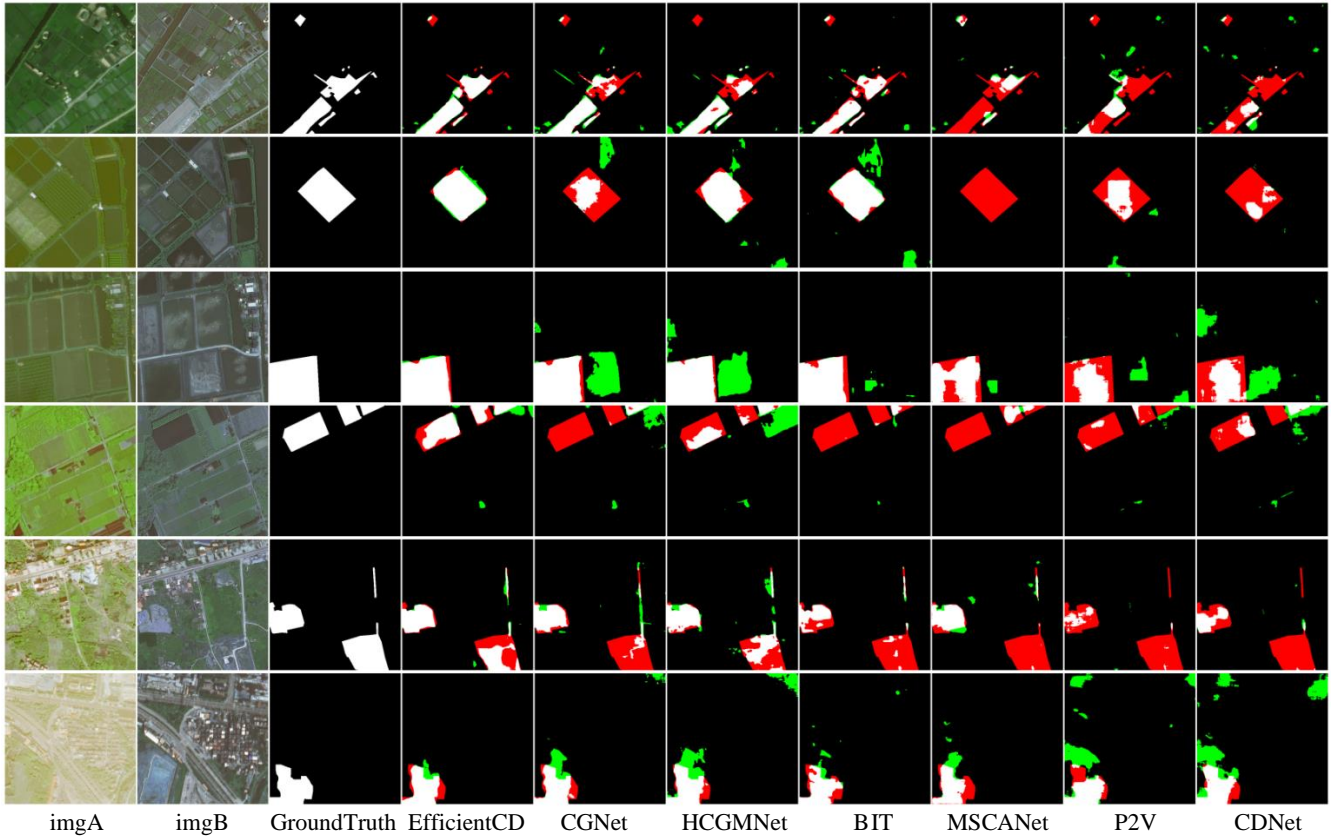

| imgA | imgB | GroundTruth | EfficientCD | CGNet | HCGMNet | BIT | MSCANet | P2V | CDNet |

**Fig. 5.** Visualization results in CLCD dataset

TABLE II
QUANTITATIVE RESULTS ON THE CLCD DATASET

| Model | OA | IoU | F1 | Rec | Prec |
|---|---|---|---|---|---|
| CDNet | 95.27 | 48.31 | 65.15 | 59.43 | 72.09 |
| P2V | 95.84 | 54.10 | 70.22 | 65.93 | 75.11 |
| MSCANet | 96.05 | 55.83 | 71.65 | 67.07 | 76.91 |
| BIT | 96.46 | 58.36 | 73.71 | 66.63 | 82.47 |
| DSIFN | 96.55 | 59.42 | 74.54 | 67.86 | 82.69 |
| HCGMNet | 96.27 | 59.58 | 74.67 | 73.92 | 75.44 |
| GaMPF | 96.68 | 60.52 | 75.40 | 68.36 | **84.06** |
| AMTNet | -- | 62.35 | 76.81 | 75.06 | 78.64 |
| CGNet | 96.82 | 62.67 | 77.05 | 71.71 | 83.25 |
| EfficientCD(B5) | **96.98** | **65.14** | **78.89** | **75.83** | 82.21 |

### B. Implementation Detail

For the implementation of EfficientCD, our framework was constructed using the PyTorch programming environment and augmented by the mmsegmentation library. The experimental setup was hosted on an Ubuntu system, equipped with NVIDIA GeForce RTX 3090 GPUs to facilitate accelerated model training. The EfficientCD and other comparison methods are trained with a batch size of 12 using distributed training on 2 GPUs. In terms of data augmentation, we used three methods: RandomRotate, RandomFlip and PhotoMetricDistortion for data enhancement. In terms of model optimization, the AdamW optimizer was utilized, configured with a learning rate set at 0.0003 and a weight decay parameter of 0.01. Throughout the experimental phase, we continuously monitored the mIoU metric on the validation set, earmarking the best-performing model for subsequent final evaluation.

### C. Evaluation Metrics

To assess our algorithm's effectiveness, we focus on metrics such as precision, recall, overall accuracy, F1-score, and Intersection over Union (IoU) for evaluating change detection. High precision reduces false positives, while high recall minimizes missed detections. These metrics are comprehensive indicators where higher values signify better performance. Their calculation formula is as follows:

$$\text{IoU} = \frac{\text{TP}}{\text{TP} + \text{FN} + \text{FP}} \quad (6)$$

$$\text{Prec} = \frac{\text{TP}}{\text{TP} + \text{FP}} \quad (7)$$

$$\text{Rec} = \frac{\text{TP}}{\text{TP} + \text{FN}} \quad (8)$$

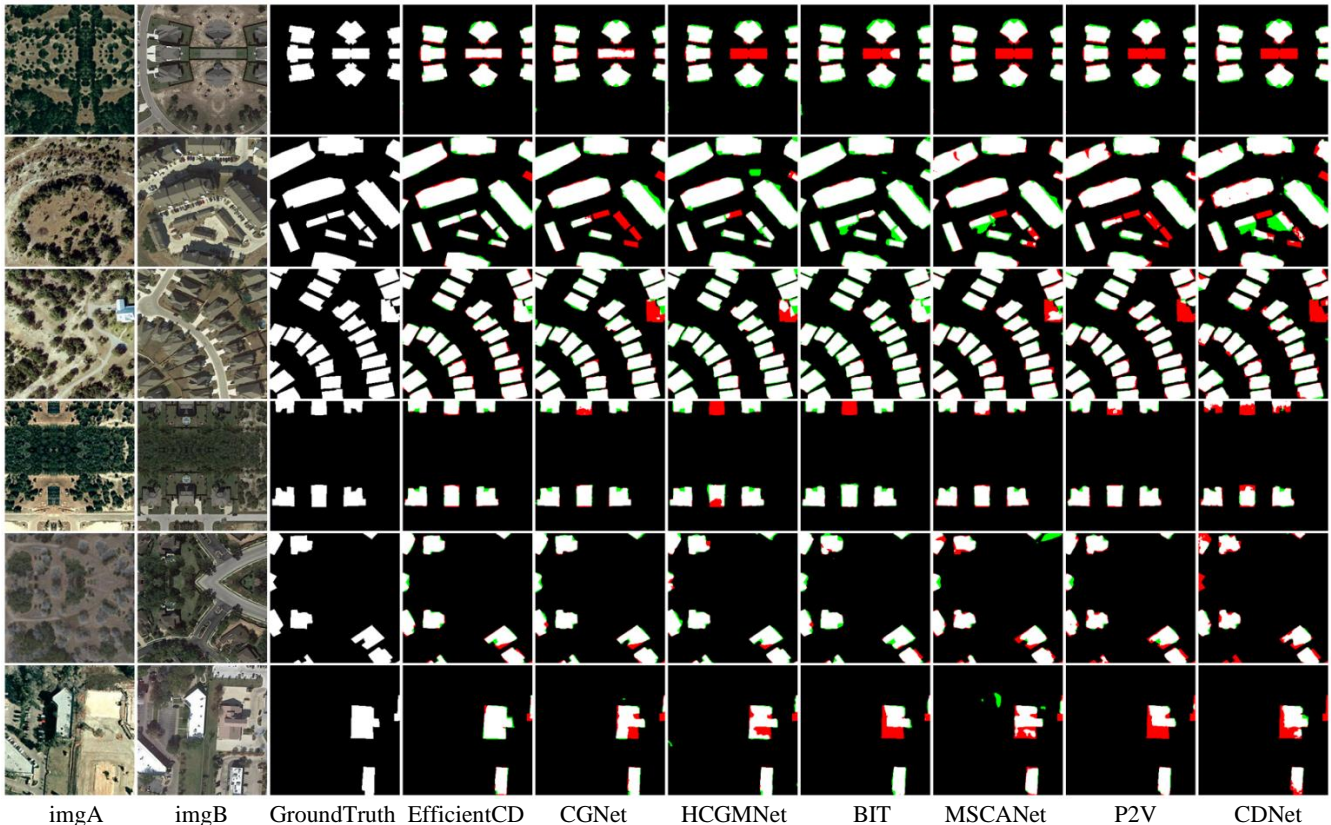

imgA    imgB    GroundTruth    EfficientCD    CGNet    HCGMNet    BIT    MSCANet    P2V    CDNet

**Fig. 6.** Visualization results in LEVIR-CD dataset

$$F1 = 2\frac{P \cdot R}{P + R} \tag{9}$$

$$OA = \frac{TP + TN}{TP + TN + FN + FP} \tag{10}$$

TABLE III
QUANTITATIVE RESULTS ON THE LEVIR-CD
DATASET

| Model | OA | IoU | F1 | Rec | Prec |
|---|---|---|---|---|---|
| CDNet | 98.35 | 72.21 | 83.87 | 84.14 | 83.61 |
| BIT | 98.95 | 80.86 | 89.48 | 87.53 | 90.65 |
| MSCANet | 99.03 | 81.91 | 90.06 | 86.38 | 94.06 |
| MFATNet | 99.03 | 82.42 | 90.36 | 88.93 | 91.85 |
| ChangeFormer | 99.04 | 82.66 | 90.50 | 90.18 | 90.83 |
| P2V | 99.04 | 83.00 | 90.71 | 91.78 | 89.67 |
| AMTNet | -- | 83.08 | 90.76 | 89.71 | 91.82 |
| DMATNet | 98.25 | 84.13 | 90.75 | 89.98 | 91.56 |
| STransUNet | 99.13 | 84.19 | 91.41 | 90.55 | 92.30 |
| HCGMNet | 99.18 | 85.26 | 92.04 | **92.81** | 91.29 |
| ChangerEx | -- | 85.29 | 92.06 | 90.56 | **93.61** |
| CGNet | 99.20 | 85.40 | 92.13 | 91.93 | 92.32 |
| EfficientCD(B5) | **99.22** | **85.55** | **92.21** | 91.22 | 93.23 |

## D. Compared Methods

In view of the change detection optimization scheme mentioned in this paper, we have extensively investigated the recent state-of-the-art change detection related algorithms for comparison. We classified these algorithms as follows:

CNN-based: CDNet [46], P2V [45], DSAMNet [43], GaMPF [47], ISNet [48], L-UNet [49]

Transformer-based: ChangeFormer [50], BIT [41], STransUNet [24], MSCANet [44], DMATNet [51]

Feature fusion-based: SSANet [52], DSIFN [53], SNUNet [54], SGSLN [55], CGNet [56], HCGMNet [57], AMTNet [58], MFATNet [59], DARNet [60]

## E. Quantify analysis and visualize results

As shown in Tables II-V, we extensively evaluated EfficientCD on four datasets: CLCD, LEVIR-CD, WHUCD, and SYSU-CD. EfficientCD achieved state-of-the-art performance across all datasets according to the main metrics. The symbol '--' denotes missing data from the original papers. For these cases, we retrained some of the models to obtain enhanced accuracy. Since these are binary change detection tasks, Intersection-over-Union (IoU) on the foreground change class serves as the primary indicator. A detailed comparison of performance across four datasets unequivocally demonstrates the superiority of the EfficientCD algorithm in critical evaluation metrics. Notably, in terms of the IoU metric, EfficientCD not only achieved the highest scores on each dataset but also displayed significant advantages: on the CLCD dataset, EfficientCD's IoU stood at 65.14%, surpassing the next highest, CGNet (62.67%), by approximately 2.47 percentage points; on the LEVIR-CD dataset, its IoU reached 85.55%, slightly above CGNet (85.40%) by 0.15 percentage points; on the SYSU-CD dataset, EfficientCD's IoU soared to 71.53%,

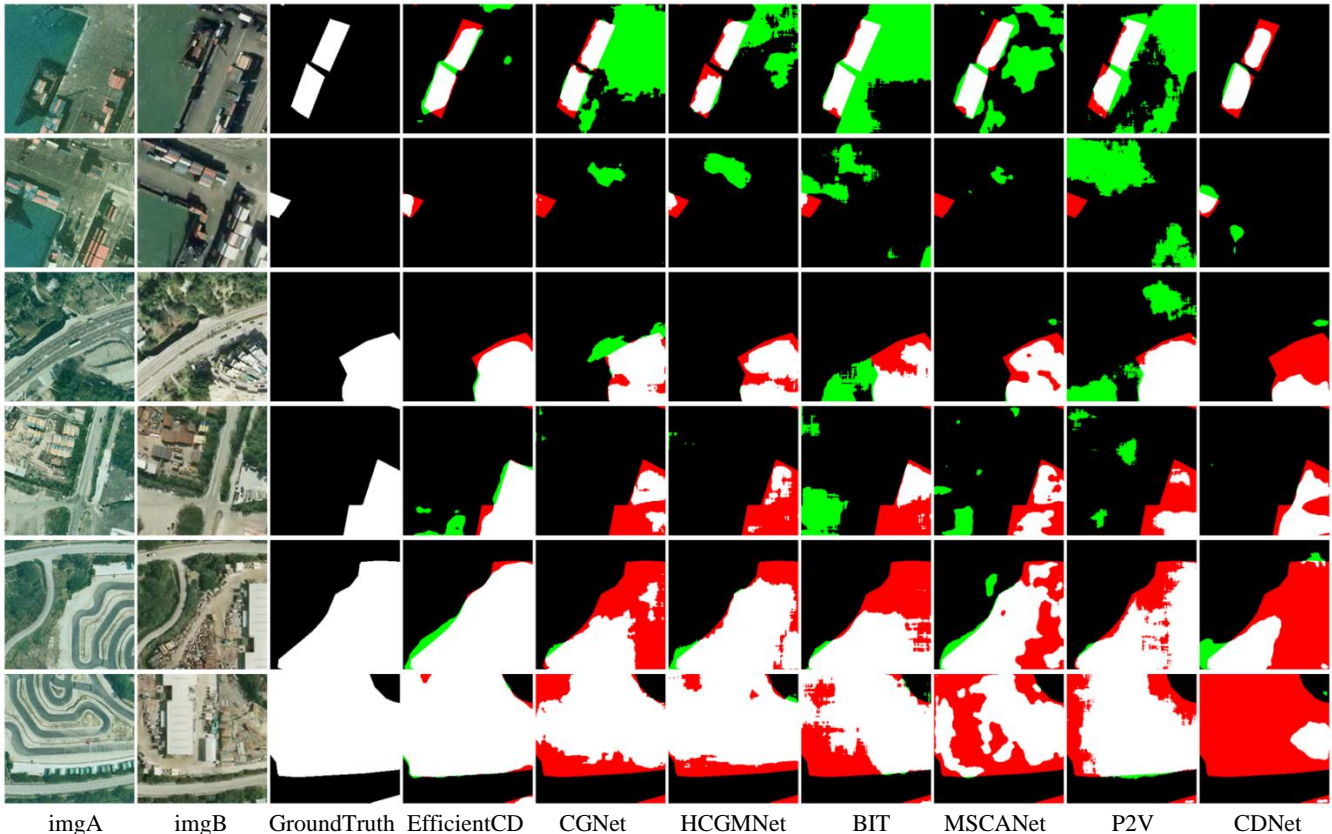

imgA  imgB  GroundTruth  EfficientCD  CGNet  HCGMNet  BIT  MSCANet  P2V  CDNet

**Fig. 7.** Visualization results in SYSU-CD dataset

TABLE IV
QUANTITATIVE RESULTS ON THE SYSU-CD DATASET

| Model | OA | IoU | F1 | Rec | Prec |
|---|---|---|---|---|---|
| MSCANet | 89.99 | 63.04 | 77.33 | 72.40 | 82.99 |
| CDNet | 89.90 | 64.34 | 78.30 | 77.29 | 79.34 |
| DSAMNet | -- | 64.18 | 78.18 | 81.86 | 74.81 |
| ISNet | 90.01 | 64.44 | 78.29 | 80.27 | 76.41 |
| BIT | 90.57 | 64.81 | 78.65 | 73.68 | 84.34 |
| SNUNet | 90.79 | 66.02 | 79.54 | 75.87 | 83.58 |
| L-UNet | 90.58 | 66.15 | 79.63 | 78.08 | 81.24 |
| P2V | 90.49 | 66.29 | 79.73 | 79.29 | 80.17 |
| HCGMNet | 91.12 | 66.33 | 79.76 | 74.15 | 86.28 |
| CGNet | 91.19 | 66.55 | 79.92 | 74.37 | 86.37 |
| DARNet | 91.26 | 68.10 | 81.03 | 79.11 | 83.04 |
| SSANet | -- | 68.18 | 81.08 | 79.73 | 82.48 |
| SGSLN | -- | 71.05 | 83.07 | **81.45** | 84.76 |
| EfficientCD(B5) | **92.46** | **71.53** | **83.40** | 80.27 | **86.78** |

exceeding the runner-up, SSANet (68.18%), by 3.35 percentage points; and on the WHUCD dataset, EfficientCD's IoU was 90.71%, overtaking CGNet (90.41%) by 0.3 percentage points. These comparative data highlights EfficientCD's efficiency and accuracy in precisely identifying change areas in remote sensing imagery, thoroughly establishing its advanced status and superiority in the field of remote sensing image change detection.

In Figures 5 through 8, we provide visualizations of our test results for CLCD, LEVIR-CD, SYSU-CD, and WHUCD. We selected models with the highest IoU scores for EfficientCD and compared them with classical algorithms from recent years. In these visualizations, True Positives (TP) are denoted by white pixels, True Negatives (TN) by black pixels, False Positives (FP) by green pixels, and False Negatives (FN) by red pixels. Our visual results unequivocally demonstrate that EfficientCD achieves excellent detection performance across diverse datasets and application scenarios, closely aligning with ground truth annotations.

*F. Ablation study*

The ablation study focused on the Intersection over Union (IoU) index across four datasets—LEVIR-CD, SYSU-CD, CLCD, and WHUCD—reveals the significant impact of incorporating ChangeFPN and the Decoder into the EfficientCD algorithm. In TABLE VI, we did not use ChangeFPN and the layer-by-layer decoder combined with Euclidean distance in the baseline. Firstly, we constructed a feature map pyramid based on the EfficientNet-B5 backbone, then fused the feature map pyramid through concatenation, and used FPN to process the fused features of the dual time phases. Finally, the processed feature maps were upsampled and restored into prediction results through hierarchical upsampling fusion. For the ablation of ChangeFPN, we removed the step of exchanging hierarchical features of dual time phases in the complete network of EfficientCD. For the ablation of the

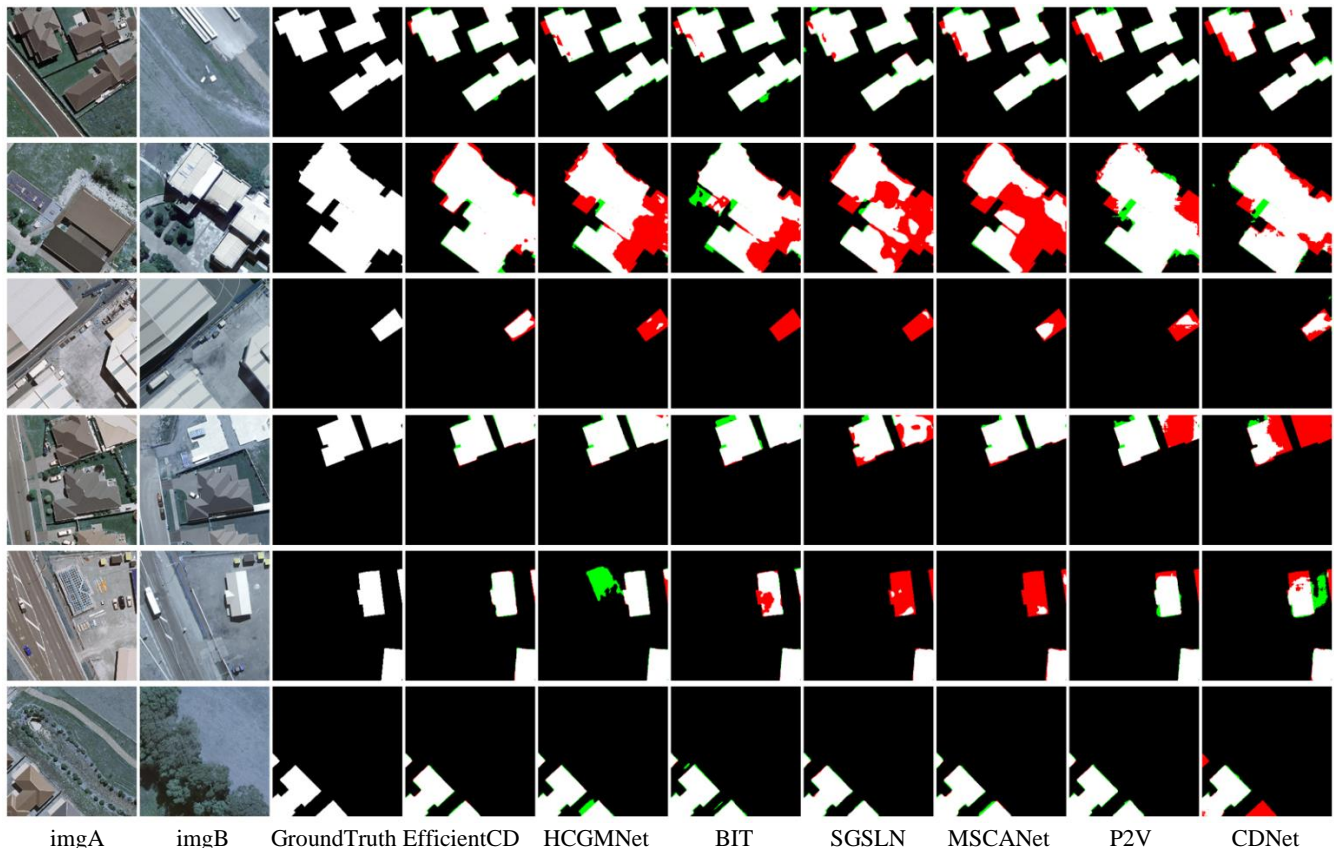

imgA imgB GroundTruth EfficientCD HCGMNet BIT SGSLN MSCANet P2V CDNet

**Fig. 8.** Visualization results in WHUCD dataset

decoder, we modified the Decoder part of EfficientCD to be consistent with the baseline. The ablation results not only demonstrate the individual contributions of ChangeFPN and Decoder to the algorithm's enhanced accuracy but also suggests a synergistic effect between them, propelling the EfficientCD algorithm to achieve its highest performance levels. This ablation study conclusively illustrates the pivotal roles that ChangeFPN and the Decoder play in improving the precision of remote sensing image change detection, underscoring their indispensable contributions to the overall effectiveness of the EfficientCD algorithm.

## V. DISCUSSION

### A. EfficientNet in Change Detection

In this paper, we employed the EfficientNet backbone network as the feature extraction module for our change detection algorithm. In binary change detection tasks, there are generally two considerations for the feature extraction part. On one hand, the change detection algorithm can be optimized by enhancing the feature extraction capability of the backbone algorithm, such as BIT, SwinUNet, etc. On the other hand, considering the model's focus in binary change detection tasks is on the change areas, we utilized a more lightweight model for feature extraction. This approach not only reduces the model's parameter count, leading to faster inference speed, but also ensures that the lightweight nature of the feature extraction backbone helps prevent overfitting.

Furthermore, to test the performance of different depths of EfficientNet models on various change detection datasets, we conducted tests on EfficientNet models from B0 to B5, and the test results are shown in the TABLE VI. EfficientNet-B6 and EfficientNet-B7 were not included in the comparative experiments due to the lack of pretrained models. From the results, it is evident that models based on the EfficientNet backbone can achieve relatively good results in change detection tasks. Additionally, the accuracy of the model on different datasets does not show a positive correlation with the depth of the EfficientNet model.

TABLE V
QUANTITATIVE RESULTS ON THE WHUCD DATASET

| Model | OA | IoU | F1 | Rec | Prec |
|---|---|---|---|---|---|
| CDNet | 98.96 | 79.59 | 88.63 | 87.24 | 90.07 |
| SNUNet | 99.24 | 84.69 | 91.71 | 90.77 | 92.67 |
| P2V | 99.31 | 85.91 | 92.42 | 90.93 | 93.97 |
| DSIFN | 99.34 | 86.36 | 92.68 | 90.20 | 95.30 |
| MSCANet | 99.36 | 86.65 | 92.85 | 89.98 | 95.90 |
| SGSLN | 99.38 | 87.47 | 93.32 | 92.91 | 93.72 |
| BIT | 99.43 | 88.22 | 93.74 | 92.00 | 95.56 |
| HCGMNet | 99.52 | 90.10 | 94.79 | **95.31** | 94.27 |
| CGNet | 99.54 | 90.41 | 94.96 | 94.61 | 95.32 |
| EfficientCD(B5) | **99.55** | **90.71** | **95.13** | 94.19 | **96.08** |

EfficientNet-B5, chosen as the backbone for the EfficientCD algorithm, demonstrates exceptional performance across various datasets, particularly excelling in SYSU-CD and LEVIR-CD with its superior IoU scores, indicating its effectiveness in complex remote sensing image change detection tasks. Despite its relatively higher parameter count and computational complexity, EfficientNet-B5 maintains a commendable balance between efficiency and precision when compared to other high-performance models like the Swin Transformer and ConvNeXt series. This balance showcases the optimized network design of EfficientNet-B5, which, while slightly increasing in parameters, achieves higher accuracy across multiple datasets. The model's adeptness at maintaining high precision with reasonable computational demands makes EfficientNet-B5 an ideal choice for high-accuracy remote sensing change detection, ensuring EfficientCD's outstanding performance across diverse datasets by leveraging an optimal trade-off between accuracy and computational efficiency.

TABLE VI
ABLATION STUDY IN IOU INDEX

| Model | ChangeFPN | Decoder | LEVIR-CD | SYSU-CD | CLCD | WHUCD |
|---|---|---|---|---|---|---|
| EfficientCD | × | × | 84.32 | 64.43 | 61.41 | 88.27 |
| | √ | × | 84.96 | 68.40 | 62.82 | 90.32 |
| | × | √ | 85.31 | 66.94 | 62.38 | 89.20 |
| | √ | √ | 85.55 | 71.53 | 65.14 | 90.71 |

TABLE VII
ABLATION STUDY IN IOU INDEX

| Backbone | Params(M) | FLOPs(G) | LEVIR-CD | SYSU-CD | WHUCD | CLCD |
|---|---|---|---|---|---|---|
| EfficientNet-B0 | 48.21 | 45.24 | 84.69 | 68.92 | 90.22 | 65.30 |
| EfficientNet-B1 | 50.71 | 45.74 | 84.82 | 69.58 | 90.06 | 61.52 |
| EfficientNet-B2 | 52.05 | 45.96 | 84.74 | 68.56 | 90.04 | 63.09 |
| EfficientNet-B3 | 55.19 | 46.77 | 84.54 | 68.02 | 90.17 | 62.38 |
| EfficientNet-B4 | 62.34 | 48.18 | 84.34 | 67.58 | 89.71 | 62.38 |
| EfficientNet-B5 | 73.43 | 50.42 | 85.55 | 71.53 | 90.71 | 64.61 |
| ResNet18 | 48.84 | 48.30 | 83.80 | 64.65 | 80.88 | 50.91 |
| ResNet34 | 58.95 | 53.12 | 82.69 | 64.97 | 82.13 | 50.94 |
| ResNet50 | 61.91 | 54.66 | 83.35 | 65.85 | 82.83 | 49.44 |
| ResNest14d | 46.96 | 51.14 | 83.80 | 66.47 | 86.31 | 55.57 |
| ResNest26d | 53.42 | 53.44 | 81.39 | 66.70 | 81.05 | 53.78 |
| ResNest50d | 63.84 | 58.03 | 73.17 | 62.74 | 72.30 | 50.65 |
| Swin_Tiny | 62.63 | 42.41 | 85.27 | 68.12 | 90.13 | 63.59 |
| Swin_Small | 83.95 | 50.93 | 85.64 | 68.28 | 90.85 | 64.65 |
| Swin_Base | 122.00 | 64.37 | 85.62 | 68.38 | 88.91 | 63.19 |
| ConvNeXt_Nano | 50.00 | 49.99 | 84.85 | 66.14 | 88.91 | 58.14 |
| ConvNeXt_Small | 62.93 | 55.29 | 84.69 | 65.25 | 89.08 | 59.21 |
| ConvNeXt_Tiny | 84.57 | 66.35 | 84.92 | 64.84 | 88.76 | 58.32 |

*B. Feature Exchange in Change Detection*

As shown in the experimental section above, we exchanged feature maps within the bi-temporal feature pyramid for change detection tasks. Without increasing the number of parameters, ChangeFPN has facilitated change detection tasks. Referring to existing research [38], we found that performing spatial or channel exchanges on bi-temporal feature maps in change detection tasks can optimize the model's detection results and enhance the its generalization. In the feature pyramid of deep learning models, features at different levels are extracted from the original image. The exchange of information between different levels of feature maps helps the model better understand image features and learn targets of different scales within the image. For this reason, we performed feature exchanges based on bi-temporal feature maps. On one hand, for bi-temporal images, ChangeFPN facilitates information interaction between features of images from different times, enhancing the representational capability of the change detection model. On the other hand, for single-temporal images, the model incorporates feature maps from another time phase, thereby supplementing the single-temporal feature extraction component with additional information.

## VI. CONCLUSION

In this paper, we designed a multi-layer feature pyramid to extract bi-temporal features based on the structural characteristics of EfficientNet. Simultaneously, without increasing additional parameter quantities, we designed ChangeFPN suitable for change detection tasks based on the FPN structure, effectively enhancing the information exchange between bi-temporal image features. Finally, in the decoder part, we combined Euclidean distance calculation to design a layer-by-layer decoder, intuitively demonstrating the differences in bi-temporal feature maps. The experimental results demonstrate that EfficientCD performs excellently on multiple standard datasets, showcasing its effectiveness in handling complex scenes and large-scale data. Additionally, the ChangeFPN proposed by us exhibits high flexibility and scalability, which can be flexibly added to change detection algorithms based on the BiEncoder-Neck-Decoder architecture. In summary, EfficientCD effectively promotes the effectiveness of change detection tasks, providing insights for future research on change detection algorithms.

ACKNOWLEDGMENT

This work was supported by the Key Research and Development Plan of Guangxi Zhuang Autonomous Region: 2023AB26007 and the National Natural Science Foundation of China (NSFC): 41971352. The numerical calculations in this paper have been done on the supercomputing system in the Supercomputing Center of Wuhan University.